\documentclass[letterpaper,10pt,conference]{ieeeconf} 
\IEEEoverridecommandlockouts  
\usepackage{url}
\usepackage{xcolor}
\usepackage{times}
\usepackage{epsfig}
\usepackage{graphicx}
\usepackage{amsmath}
\usepackage{amssymb}
\usepackage{bbm,dsfont}
\let\labelindent\relax
\usepackage{enumitem}

\usepackage{times,epsfig,graphicx}
\usepackage{algorithm,algorithmic}
\usepackage{amssymb,amsmath}
\usepackage{siunitx}

\makeatletter

\DeclareRobustCommand\onedot{\futurelet\@let@token\@onedot}
\def\@onedot{\ifx\@let@token.\else.\null\fi\xspace}

\def\be{\begin{equation}}
\def\ee{\end{equation}}
\def\bea{\begin{eqnarray}}
\def\eea{\end{eqnarray}}
\def\ben{\begin{eqnarray*}}
\def\een{\end{eqnarray*}}

\def\bi{\begin{itemize}}
\def\ei{\end{itemize}}

\newcommand{\bt}[1]{\begin{tabular}{#1}}
\newcommand{\et}{\end{tabular}}
\newcommand{\ba}[1]{\begin{array}{#1}}
\newcommand{\ea}{\end{array}}

\def\<{\langle}
\def\>{\rangle}

\newcommand{\hide}[1]{}

\newcommand{\bfx}{{\bf x}}

\newcommand{\ra}{{\rightarrow}}

\DeclareMathOperator*{\E}{\mathbb{E}}

\newcommand{\blue}[1]{}

\newtheorem{assumption}{Assumption}

\begin{document}

\title{\LARGE Universal Domain Adaptation in Ordinal Regression}

\author{Chidlovskii Boris\\
Naver Labs Europe\\
ch. Maupertuis 6, Meylan\\
France \\
\and
Assem Sadek\\
Naver Labs Europe\\
ch. Maupertuis 6, Meylan\\
France \\
\and
Christian Wolf\\
INSA-Lyon, LIRIS\\
UMR CNRS 5205, Villeurbanne\\
France
}

\maketitle

\begin{abstract}
We address the problem of universal domain adaptation (UDA) in ordinal regression (OR), which attempts to solve classification problems in which labels are not independent, but follow a natural order. We show that the UDA techniques developed for classification and based on the clustering assumption, under-perform in OR settings. 
We propose a method that complements the OR classifier with an auxiliary task of order learning, which plays the double role of discriminating between common and private instances, and expanding class labels to the private target images via ranking. Combined with adversarial domain discrimination, our model is able to address the closed set, partial and open set configurations. We evaluate our method 
on three face age estimation datasets, and show that it outperforms the baseline methods.
\end{abstract}

\section{Introduction}
\label{sec:introduction}




Domain adaptation techniques were introduced to address the domain shift between source and target domains~\cite{TorralbaCVPR11Unbiased,zhao20review}. Domain adaptation (DA) is a form of transfer learning that aims to learn a model from a labeled source domain that can generalize well to a different (but related) unlabeled or sparsely labeled target domain. 
Domain adaptation has demonstrated a significant success in various applications, including image classification~\cite{HoffmanX13Efficient}, 
semantic segmentation~\cite{zhang17curriculum}, object recognition~\cite{LongCVPR14Transfer}  object detection, 3D point cloud segmentation~\cite{zhao20review}, etc.
These successes are due to the same semantic space shared by source and target domains. Common classes implicitly structure the output space, where separation between two classes in the source can be transferred to the target. Moreover this knowledge makes possible totally {\it unsupervised} domain adaptation~\cite{HoffmanX13Efficient,LongICML15Learning}.

More challenging problems of {\it partial} and {\it open set} domain adaptation have been recently addressed~\cite{cao2018-eccv,Saito_2018_ECCV}. In {\it universal domain adaptation}(UDA)~\cite{saito20universal,you2019universal}, source and target domains share a set of common classes but each domain may have its private classes. The task is to identify source and target images in common classes to apply domain adaptation.
Existing UDA methods 
count on entropy-based and uncertainty criteria to weigh down images estimated as being in private source classes~\cite{cao2019learning,liang20balanced} or target clusters~\cite{hu20unsupervised,li2020enhanced,pan2020exploring}
and mark all private target images as “unknown”.

In this paper we investigate the UDA problem in {\it ordinal regression} aiming to solve a classiﬁcation problem where classes are not independent but follow a natural order.
Ordinal regression (OR) is an important research topic in machine learning~\cite{diaz19soft,fu18,liu18constrained,niu16}. 
Its applications include age estimation~\cite{li19bridgenet,lim20order,niu16}, depth estimation~\cite{fu18}, human preferences~\cite{diaz19soft} and ratings~\cite{gutierrez16ordinal}. Moreover, any 1D regression task is convertable in OR by discretizing the target variable~\cite{chu05gausian}.
To take into account the class order, multiple shallow and deep methods have been developed for OR~\cite{cao2020rankconsistent,diaz19soft,gutierrez16ordinal,niu16,pang20}. In this paper we build on the recent state of the art method of consistent rank logits (CoRaL)~\cite{cao2020rankconsistent}.

Domain shift in OR faces the same challenges as in classification. Transferring source OR classifier to target domain requires learning domain invariant representations. If source and target domains share the same classes, multiple well known solutions~\cite{ganin2016,hoffman18cycada,LongICML15Learning} 
can be adopted by ignoring the class order or by replacing the cross entropy loss in source classifier with the OR loss~\cite{cortes2011domain}.

However, domain shift in OR usually comprises the output shift (also called {\it category shift}). Consider the example of domain shift in human face age estimation. UTKFace and AFAD are two popular datasets for training the age estimation models~\cite{cao2020rankconsistent,niu16}. AFAD 
is a collection of Asian face images aged between 15 and 40 years old; UTKFace is a universal face dataset with the age range from 1 to 80 years. 
The domain shift combines the input shift (faces of different human races) and output shift (ages). 
A model trained on UTKFace and adopted to AFAD should be able to constrain estimations to the target age range. Inversely, a model trained on AFAD and adopted to UTKFace should detect extra classes and, if possible, make predictions beyond the source age range.  

Existing UDA methods for classification~\cite{cao2018-eccv,cao2019learning,liang20balanced,pan2020exploring,saito20universal,you2019universal} perform sub-optimally under output shift in OR, for two reasons. First, they assume class independence and follow an entropy minimization approach to enforce the decision boundary to pass through low-density area. The uncertainty of class predictions serves to detect outlier source classes in partial DA~\cite{cao2018-eccv};
low confidence entropy helps detect source and target private images in open set DA~\cite{you2019universal};
neighbour clustering exploits the cluster structure of the target domain for cluster matching~\cite{saito20universal}.



However the clustering assumption is often invalidated in OR. By their nature, ordinal data span a low-dimensional manifold where classes are assigned via thresholding~\cite{gutierrez16ordinal,liu11ordinal}. This can confuse the entropy-based criteria for separating common and private images and lead to domain misalignment and negative transfer.
 
Second, as classes are treated independently, existing UDA methods are prone to making boundary decision {\it order inconsistent} where classes marked as common mix up with classes marked as private. 

\begin{figure}[t]
\centering
\includegraphics[width=0.7\columnwidth]{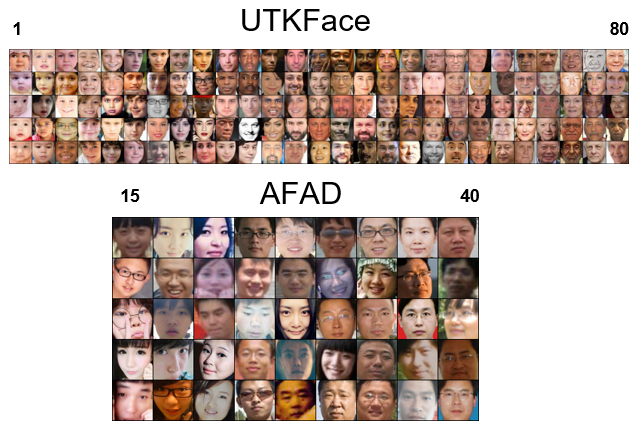}
\caption{Input shift (faces of different human races) and output shift (age range) in UTKFace and AFAD datasets for facial age estimation.}
\label{fig:2utk-afad}
\end{figure}

\begin{figure}[t]
\centering
\includegraphics[width=\columnwidth]{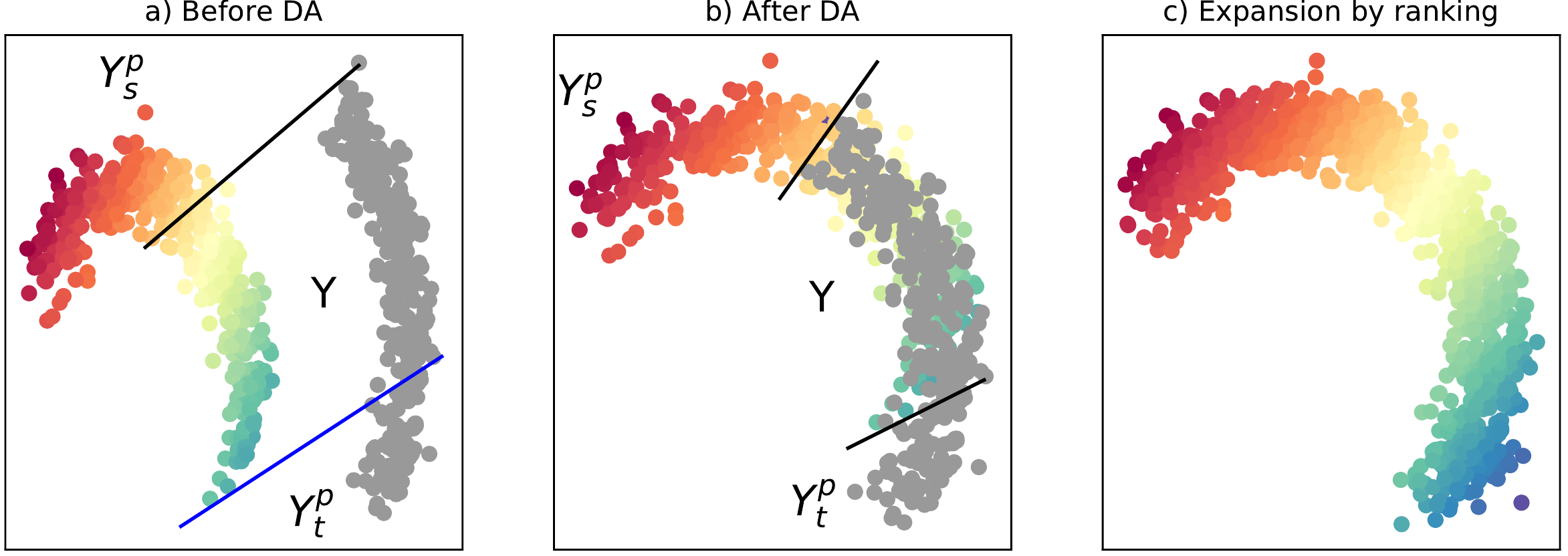}
\caption{Universal domain adaptation on source/target manifolds: $Y$ is common, $Y_s^p$ and $Y_t^p$ are private source and target parts (see Section~\ref{sec:uda}). (a) domain misalignment before adaptation; (b) after manifold alignment on common classes; (c) extending labels to the target private classes.}
\label{fig:2manifolds}
\end{figure}

If trained separately, source and target domains form two separate manifolds due to their domain specific features (see Figure~\ref{fig:2manifolds}.a). 
Learning domain-invariant representations~\cite{ganin2016,LongICML15Learning} 
enables forming one unique manifold (see~Figure~\ref{fig:2manifolds}.b).
However, to avoid the misalignment, the DA method should accurately separate common and private classes and train the domain discriminator on the common classes only. 


\paragraph{Auxiliary task of order learning.}
To address the partial and open set DA in OR, we introduce an auxiliary task of order learning~\cite{lim20order}. Order learning performs pairwise comparison between images, instead of directly estimating the class of an image. The advantage of such a complementary representation is two-fold.
First, the order model aims to replace the cluster assumption when detecting the boundary between common and private classes. 

Second, it can benefit from the aligned manifold structure and predict the relative position of private target images w.r.t. the common class set $Y$. Predicting exact classes requires some target labels~\cite{chidlovskii2020adversarial,lim20order} and is beyond the unsupervised domain adaptation setting. However, we can assume homogeneity of relative positions of images in the common and private classes.
Under this assumption, the order model can compare target private images pairwisely, 
transform these comparisons in the ranking and convert the ranks into classes (see~Figure~\ref{fig:2manifolds}.c). In other words, we explore the ordinal data structure to assign true classes to private images.

Trained on source pairs, the order model can suffer from the negative transfer, just as the OR does. We propose a deep architecture for UDA which jointly learns OR, the order and domain invariant features through an adversarial domain discrimination. 
We then leverage the learned order relationships to rank private target images.

The contribution of this paper is three-fold.
First, we introduce a new UDA solution, that focuses on OR tasks. We show that the existing UDA methods that process classes independently under-perform on OR in the presence of common and private classes. Second, we propose a solution based on order learning to address four UDA configurations: {\it same classes} (SC), {\it partial} (PA), {\it open set} (OS) and {\it partial/open set} (OSPA). We show how to train the order model to detect the boundary between common and private classes and weigh down private instances in domain discrimination.
Third, we show that the order model permits to replace the ``\emph{unknown}'' class assigned to private target instances with a ranking and, under a mild assumption, to convert these ranks into true target classes.

\section{Related work} 
\label{sec:sota}

There exist multiple families of solutions for domain adaptation~\cite{csurka17,zhao20review}. 
Feature-level strategies focus on learning domain invariant data representations mainly by minimizing different domain shift measures~\cite{SunX15Return}. 
Discrepancy-based methods explicitly measure the discrepancy between the source and target domains on corresponding activation layers of the two network streams~\cite{LongICML15Learning}. 
Optimal transport models try to align the representations of the source and target domains
via the optimal distribution matching~\cite{courty17optimal,li2020enhanced} and domain structure exploration~\cite{xu2020reliable}.

Domain adversarial training of neural networks is another popular approach to learn domain invariant and task discriminative representations~\cite{ganin2016}. Adversarial discriminative models employ an adversarial objective with respect to a domain discriminator to encourage domain confusion~\cite{tzeng17adversarial}. 
This approach has multiple variants, some of which also exploit class-specific and group-specific domain recognition components ~\cite{Li2019_ontarget,Saito_2018_ECCV}. 




\paragraph{Universal Domain Adaptation.}
Recent works revised the main assumption of the source and target sharing the same class set and study more realistic scenarios~\cite{cao2018-eccv,cao2019learning,saito20universal,Saito_2018_ECCV,you2019universal}. In {\it partial DA}, the target is allowed to cover only a subset of the source class set. In this case the adaptation process should be adjusted so that the samples with not-shared labels would not influence the learned model. Common techniques consist in adding a re-weight source sample strategy to a standard DA approach~\cite{cao2018-eccv,zhang2018cvpr}. Alternative solutions leverage two separate deep classifiers and their prediction inconsistency on feature norm matching~\cite{xu18nonparametric}, or include adversarial alignment and adaptive uncertainty suppression~\cite{liang20balanced}.

The {\it open set} scenario is not trivial since the target samples in unknown class are not expected to align with the source. This problem is addressed by measuring the entropy of class  predictions~\cite{you2019universal} or by augmenting the domain adaptation with class-agnostic clusters in target domain~\cite{pan2020exploring}. 
\cite{hu20unsupervised} proposed a method to model the synchronization relationship among the local distribution pieces and global distribution, aiming for more precise domain-invariant features in hierarchical manner.
The sample transfer scoring scheme in~\cite{lifshitz21sample} 
couples target pseudo-labeling with careful sample selection to ensure class diversity in a batch.

A new UDA framework has been proposed in~\cite{saito20universal}. 
It relies on a neighborhood clustering to learn the structure of the target domain in a self-supervised way.  It also uses entropy-based feature alignment and rejection to align target features with the source, or reject them as ``\textit{unknown}''. 


All mentioned methods address the classification tasks.
Domain adaptation in regression is less studied~\cite{zhao20review} and represented by eye gaze~\cite{shrivastava17learning} and hand pose estimation~\cite{chen16robust,NeverovaCVIU2017}. In both tasks, a model trained on synthetically generated images is transferred to real images by adjusting the loss function or by reweighting source images. The process of generating synthetic images ensures the same output space across the domains and therefore prevents any output shift. In \cite{NeverovaCVIU2017}, this is addressed exploiting topological properties, which are invariant to the shift across domains.

\subsection{Ordinal regression}
\label{ssec:ordinal}

Ordinal regression is also called ordinal classiﬁcation
~\cite{diaz19soft,niu16,pang20}. Its goal is to predict the category of an input instance from a discrete set of classes which form a natural order. Common examples of such tasks are movie ratings, customer satisfaction surveys, age estimation, etc.~\cite{cao2020rankconsistent,fu18,gutierrez16ordinal}. 

In a sense, ordinal regression attempts to solve classiﬁcation problems in which wrong classes are not equally wrong. In age estimation, if a given person is 25 years old, estimation of his age of 30 years is less incorrect than the estimation of 40 years. 

Shallow methods framed the OR problem as classiﬁcation with a set of thresholds on the output space~\cite{gutierrez16ordinal}. 
In deep learning,~\cite{niu16} first proposed Ordinal Regression CNN by reducing ordinal regression 
with $m$ classes into $m-1$ binary classiﬁcation problems, with the $k$-th classifier predicting whether the age label of a face image exceeds rank $r_k, k = 1,\ldots,m-1$. While the binary classification approach is able to achieve state-of-the-art performance, it does not guarantee consistent predictions, such that predictions for individual binary tasks may disagree.  
This inconsistency leads to sub-optimal performance when the $m-1$ task predictions are combined to obtain the estimated age.

The consistent rank logits method (Coral)~\cite{cao2020rankconsistent}
addresses the inconsistency issue and provides a theoretical guarantee for classiﬁer consistency without increasing training complexity.
Using the binary classifier responses, the predicted class
for an input $\bfx$ is obtained by
\begin{equation}
    h(\bfx) = y_q, q = 1 + \sum_{k=1}^{m-1} f_k(\bfx),
\label{eq:or-sum}    
\end{equation}
where $f_k(\bfx) \in \{0, 1\}$ is the prediction of the $k$-th binary classiﬁer in the output layer. Classifiers $f_k, k=1..m-1$ are required to reﬂect the ordinal information and to be {\it order-monotonic}, 
$f_1(\bfx) \geq f_2(\bfx) \geq \ldots \geq f_{m-1}(\bfx)$, which guarantees consistent predictions. To achieve order-monotonicity and guarantee binary classiﬁer consistency, the $m-1$ binary tasks 
share the same intermediate layers but are assigned distinct weight parameters in the output layer~\cite{cao2020rankconsistent}.
\begin{figure}[t]
\centering
\includegraphics[width=\columnwidth]{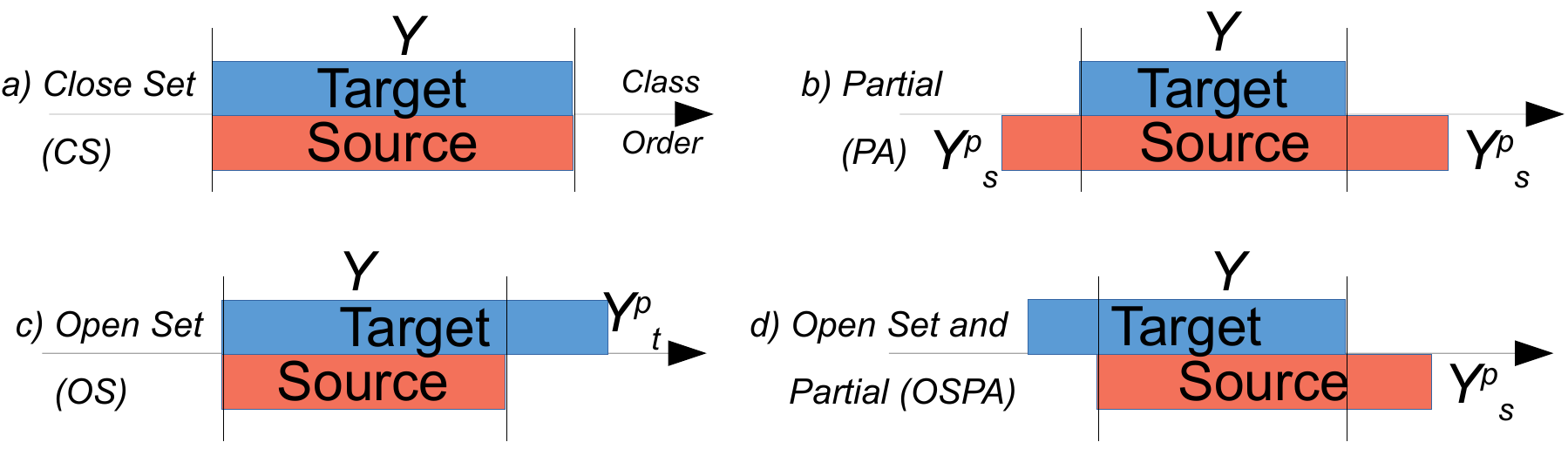}
\caption{Four UDA configurations in ordinal regression according to label space overlap: (a) close set (CS); (b) partial (PA), (c) open set (OS),(d) both partial and open set (OSPA).}
\label{fig:4cases}
\end{figure}

\section{Universal Domain Adaptation in OR} 
\label{sec:uda}
We address the problem of Universal Domain Adaptation in OR, where a source domain $D_s = \{(\bfx_s^i, y_s^i)\}$ consisting of $n_s$ labeled samples and a target domain $D_t = {(\bfx_t^i)}$ of $n_t$ unlabeled samples are provided at training. We use $Y_s$ to denote $m_s$ classes of source domain, $Y_s=(y_{s_1},y_{s_1}+1,\ldots,y_{s_1}+m_s-1)$, and $Y_t$ to denote $m_t$ classes of target domain, $Y_t=(y_{t_1},y_{t_1}+1,\ldots,y_{t_1}+m_t-1)$. 

The set of common classes shared by both domains is denoted $Y = Y_s \cap Y_t$. Private classes in source and target domains are denoted $Y_s^p = Y_s \setminus Y$ and $Y_t^p = Y_t \setminus Y$. Target data is fully unlabeled, the target class set is only used for defining the UDA problem. We define the {\it commonness} between two domains~\cite{cao2019learning} as the Jaccard distance of two class sets, $\xi = \frac{|Y_s \cap Y_t|}{|Y_s \cup Y_t|}.$

The smaller $\xi$ is, the less knowledge is shared by the domains. 
We assume that $\xi>0$ and therefore the common class set $Y$ is not empty. Figure~\ref{fig:4cases} shows four possible configurations of UDA in OR. Case (a) shows the {\it closed set} (CS) when $Y_s = Y_t = Y$; case (b) refers to {\it partial} domain adaptation (PA) when $Y_t^p = \emptyset$; case (c) refers to the {\it open set} (OS) when $Y_s^p = \emptyset$; case (d) combines the partial and open set (OSPA).
Our goal is to design a model that works well across a wide spectrum of $\xi$. It must be able to distinguish between images data coming from the common set $Y$ and private sets $Y_s^p$ and $Y_t^p$.

\begin{figure}[t]
\centering
\includegraphics[width=\columnwidth]{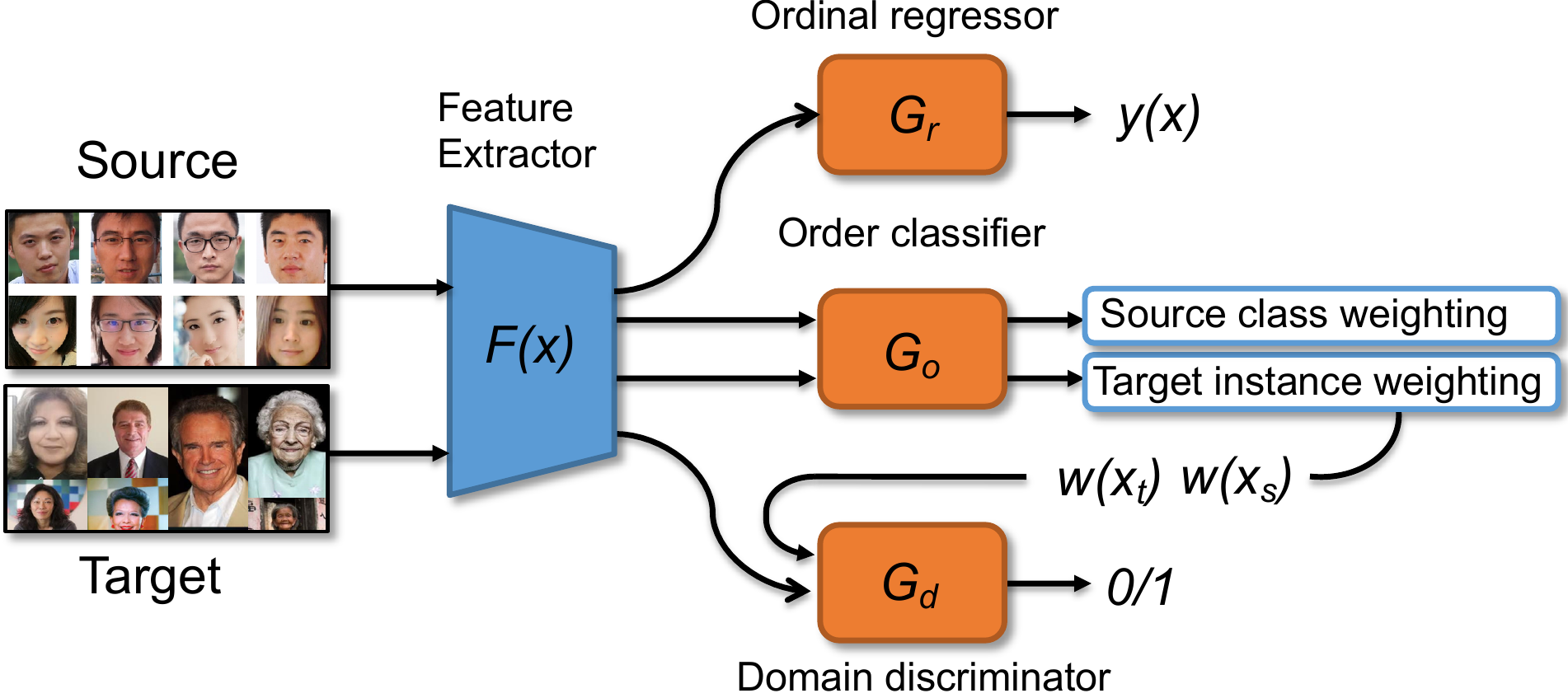}
\caption{The ORUDA model includes four main modules: feature extractor $F$, ordinal regressor $G_r$, order classifier $G_o$ and domain discriminator $G_d$. Source class weighting and target instance weighting serve to discriminate common and private classes.}
\label{fig:architecture}
\end{figure}

\subsection{UDA for OR through the manifold hypothesis}
\label{ssec:oruda}

We propose a method that trains a transferable feature extractor $f=F(\bfx)$ and an ordinal regressor $y=G_r(z)$. Trained on labeled source images and unlabeled target images, our model provides an accurate adaptation of the source OR model to the target domain, by adopting the domain-level adversarial discriminator $G_d$~\cite{cao2019learning,ganin2016,you2019universal} to reduce the discrepancy between domains and learn {\it domain-invariant} image representations $f=F(\bfx)$. 

The special UDA requirements for handling samples from private target classes $Y_t^p$ during learning, and to make predictions for them during testing, are dealt with the following assumption:
\begin{assumption}[Manifold assumption]\label{as:manifold}
~\\
\begin{enumerate}[label=(\alph*)]
\item The ordinal instances live in a low-dimensional manifold where classes form a natural order.
\item The relative difference/order between images 
on the manifold is a monotonic function invariant to the domain shift.
\end{enumerate}
\end{assumption}
Assumption~\ref{as:manifold} 
frames our learning approach. 
Assumption~\ref{as:manifold}.a is exploited during learning through a new pairwise order classifier $G_o(f_1,f_1)$, which predicts whether a pair of samples $(\bfx_1,\bfx_2)$, and in particular their domain invariant features $f_1=F(\bfx_1), f_2=F(\bfx_2)$, follows the class order 
or not. Building on this classifier, we design a procedure providing estimates on whether given target samples (whose labels are unknown) are in common or private classes, and reweigh their contribution to learning accordingly. This is detailed in Section~\ref{ssec:order}.

Assumption~\ref{as:manifold}.b is exploited during the testing. Instead of classifying all samples using the ordinal regressor $G_r$, which would be acting out of its domain in the case of samples of the private target space $Y_t^p$, we deal with these samples differently. We first mark these images as private using the order classifier, and we take decisions jointly for the full set of private samples, assigning class labels according to Assumption~\ref{as:manifold}.b; this is described in detail in Section~\ref{ssec:ranking}.

Figure~\ref{fig:architecture} illustrates the UDA model for OR. All different branches $F, G_r, G_d$ and $G_o$ are trained jointly end-to-end in a minimax optimization procedure, where the feature extractor $F$ is trained by maximizing the loss of domain discriminator $G_d$, while $G_d$ is trained by minimizing the domain discrimination loss, as described in Section~\ref{sec:arch}.

The order model is the main novelty in our method, we describe it in details in the following section. 

\subsection{Order learning}
\label{ssec:order}
We start by defining the ideal scenario, which defines an order relationship between input samples through the relationships between their classes. Building on this definition, we detect private target images and exclude them from domain discrimination by assigning low weights. For the source domain, we follow a similar approach and weigh down classes detected as being private. Our initial definition is deterministic and does not yet model any uncertainties. This definition will then be relaxed and implemented as trained network in the subsequent section.

Let $\bfx_1$ and $\bfx_2$ be two images belonging to classes in $Y_s$. Their ordering relationship is defined according to their classes as follows
\vspace*{-1mm}
\begin{equation}
\bfx_1 \prec \bfx_2 \quad {\rm if} \quad y(\bfx_1) \le y(\bfx_2) +\tau,
\label{eq:order}
\end{equation}
where $\tau$ is a threshold~\cite{lim20order}. Note that we use ‘$\prec$’ for the instance ordering and ‘$\le$’ for the class order. 

{\bf Private target instance.}
By the definition of class order, a target image $\bfx_t$ is {\it private} if it is bigger or smaller than {\it all} source images. 
Let $p(\bfx_t \prec D_s)$ denote the probability of target image $\bfx_t$ to be smaller than all source images,
\begin{equation}
p(\bfx_t \prec D_s) =\E_{\bfx_s \in D_s}\mathbbm{1}(\bfx_t \prec \bfx_s),
\label{eq:x_t}
\end{equation}
where the indicator function $\mathbbm{1}(\cdot)$ is 1 if the condition is satisfied, 0 otherwise. In the ideal case, when all order relationships $\bfx_t \prec \bfx_s$ are known, a value 1 (or 0) of $p(\bfx_t \prec D_s)$ means that $\bfx_t$ is smaller (or bigger) than the source set and therefore is private. All intermediate values of $p(\bfx_t \prec D_s)$ indicate that $\bfx_t$ is in the common set $Y$.\footnote{
We say that $\bfx_s \in Y$ if its label is in $Y$, $y(\bfx_s) \in Y$.} This binary decision can be expressed as $p(\bfx_t \in Y) =\epsilon(p(\bfx_t \prec D_s))$ where $\epsilon()$ is a binary filter
\vspace{-2mm}
\begin{equation}
\epsilon(x) = \left\{
    \begin{array}{ll}
        0 & \mbox{if } x \in \{0,1\} \\
        1 & \mbox{otherwise.}
    \end{array}
\right.
\label{eq:epsilon}
\end{equation}

\vspace{-2mm}
{\bf Private source class.}
Symmetrically, we denote $p(\bfx_s \prec D_t)$ the probability that a source image $\bfx_s$ is smaller than all target images. A source class $y_s$ is {\it private} if all images of this class are bigger or smaller that all target images, $p(y_s \in Y_s^p) = \epsilon(p(y_s \prec D_t))$, where
\begin{equation}
\begin{tabular}{lll}
$p(y_s \prec D_t)$&=&$\E_{\bfx_s, y(\bfx_s)=y_s} p(\bfx_s \prec D_t)$\\
                  &=&$\E_{\bfx_t \in D_t, \bfx_s, y(\bfx_s)=y_s} \mathbbm{1} (\bfx_s \prec \bfx_t)$
\end{tabular}
\label{eq:y_s}
\end{equation}
and $\epsilon(\cdot)$ is the binary filter (\ref{eq:epsilon}).


Eqs.~(\ref{eq:x_t})-(\ref{eq:y_s}) define three steps to estimate if a target image $\bfx_t$ or source class $y_s$ is in the common set $Y$. First, we dispose a model to estimate the order relationship for any two images $\bfx_1$ and $\bfx_2$. Second, we estimate $p(\bfx_t \prec D_s)$ and $p(y_s \prec D_t)$ using Eqs.~(\ref{eq:x_t}) and (\ref{eq:y_s}). Third, we apply the binary filter~(\ref{eq:epsilon}) to estimate $p(\bfx_t \in Y)$ and $p(y_s \in Y)$.

\begin{figure}[t]
\centering
\includegraphics[width=0.45\columnwidth]{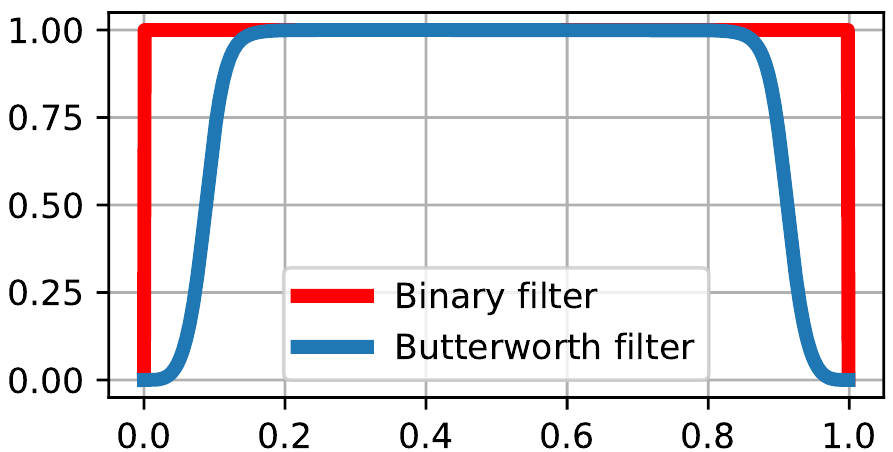}
\caption{Binary vs smoothed filters.}
\label{fig:filter}
\end{figure}

\subsection{Training the order model}
\label{ssec:s_pr}
The ideal scenario presented in the previous section assumes knowing all order relationships between source and target images. In this section we adjust the ideal scenario to the UDA setup, where we know source labels only and have no order relationships between source and target images. 

We train the order model to estimate $p(\bfx_1 \prec \bfx_2)$ using source image pairs. And we can not expect the model to be 100\% accurate as it is shown in~\cite{lim20order}. To tolerate order model errors, we first replace the binary filter (\ref{eq:epsilon}) with a smoothed filter $\tilde \epsilon(x)$ when $x$ values are close to 0 and 1. Figure~\ref{fig:filter} plots the binary filter (\ref{eq:epsilon}) and the smoothed filter implemented as a symmetric $n$-order {\it Butterworth low-pass filter} widely used in signal processing~\cite{proakis1992}. The filter is available in {\tt scipy} package and controlled by two parameters, width $w_\epsilon$ and order $n_\epsilon$.\footnote{
https://docs.scipy.org/doc/scipy/reference/generated/scipy.signal.butter.html.}
 

\begin{figure}[t]
\centering
\includegraphics[width=1\columnwidth]{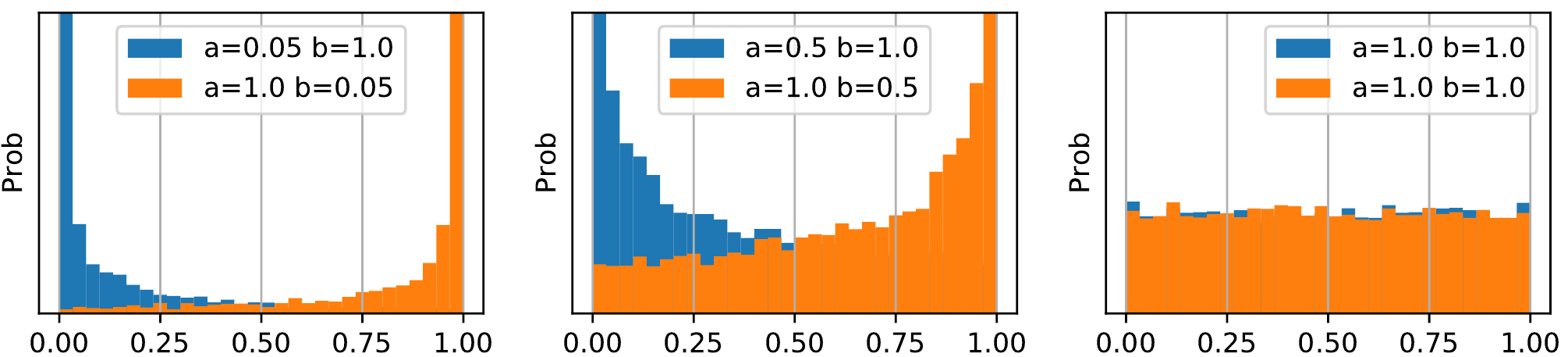}
\caption{Beta distribution functions $B(\alpha,1.0)$ and $B(1.0,\alpha)$ for curriculum learning, $\alpha=0.05, 0.5,1.0$.}
\label{fig:curr}
\end{figure}
\paragraph{Curriculum learning.}
The order model is trained on source image pairs $(\bfx_1,\bfx_2)$. Since the exhaustive enumeration of image pairs is computationally expensive, we sample source pairs. Moreover we avoid random sampling and propose a special strategy aimed at faster training. 

Curriculum learning (CL)~\cite{zhang17curriculum} proceeds by presenting easy examples to the learner before hard ones. For age estimation, easy examples are those having a clear visual clue on which face is younger. So we initially present image pairs having a big age difference. 
We start with the symmetric $\lfloor$-shape (\textcolor{blue}{blue}) and $\rfloor$-shape (\textcolor{orange}{orange}) distributions for sampling image pairs $(\bfx_1,\bfx_2)$ in Figure~\ref{fig:curr}.a. 

The sampling policy changes over epochs by presenting harder pairs with smaller class difference. The sampling distribution at epoch $ep$ is defined by Beta function, $B(\alpha,1.0)$ for image $\bfx_1$ and $B(1.0,\alpha)$ for image $\bfx_2$, where $\alpha = 1-exp(ep)$. Symmetric distributions flatten over epochs (Figure~\ref{fig:curr}.b) and tend to the uniform distribution $B(1,1)$ (Figure~\ref{fig:curr}.c) which mixes up easy and hard pairs.

\paragraph{Target images and source class in common class set $Y$.} 
The probability of a target image $\bfx_t$ to be in $Y$ is estimated as $p(\bfx_t \in Y)=\tilde \epsilon (p(\bfx_t \prec D_s))$, 
by applying the order model on randomly sampled $l_s$ source images, $p(\bfx_t \prec D_s) = \E_{\bfx_s \sim D_s} p(\bfx_t \prec \bfx_s)$. 



The probability of source class $y_s$ to be common is $p(y_s \in Y)=\tilde \epsilon (p(y_s \prec D_t))$, where
\begin{equation}
  p(y_s \prec D_t) =\E_{\bfx_s, y(\bfx_s)=y_s, \bfx_t \sim D_t} p(\bfx_s \prec \bfx_t),  
\end{equation}
is the average over all images in $y_s$
compared with randomly sampled $l_t$ target images.

\section{Optimization and training} 
\label{sec:arch}

We now define losses for three main components of the ORUDA network (see Fig~\ref{fig:architecture}): source regressor $G_r$, 
order classifier $G_o$ and domain discriminator $G_d$.
First, the OR loss for $G_r$ is defined on labeled source images,
\begin{equation}
{\cal L}_{or}(F,G_o)= \E_{(\bfx_i,y_i) \in D_s} L_{coral} (G_r(F (\bfx_i)),y_i),
\label{eq:loss_or}
\end{equation}
where $L_{coral}$ is the Coral loss defined in Section~\ref{ssec:ordinal}.

Second, the order loss for $G_o$ is defined on pairs of source images and their order relationships,
\begin{equation}
{\cal L}_{ord}(F,G_o)= \E_{\bfx_i,\bfx_j \sim D_s} 
        L_{ord} (G_o, F',\bfx_i\prec\bfx_j),
\label{eq:loss_src}
\end{equation}
where $L_{ord}$ is the cross entropy loss, $F'$ compares the two image feature vectors. Instead of feature concatenation implemented in~\cite{lim20order}, we use the vector difference, $F'=F(\bfx_i)-F(\bfx_j)$.

Third, the adversarial domain discriminator $G_d$ is trained to distinguish between feature representations of source and target images, with the adversarial loss
\begin{equation}
\begin{tabular}{ll}
${\cal L}_{d}$=
     & $\E_{\bfx_s \in D_s} w(\bfx_s) \log G_d(F(\bfx_s))$ + \\ 
     & $\E_{\bfx_t \in D_t} w(\bfx_t) \log (1-G_d(F(\bfx_t))$,
\end{tabular}
\label{eq:l_gan}
\end{equation}
where weights $w(\bfx_t)$ and $w(\bfx_s)$ for target and source images are obtained in Section~\ref{ssec:s_pr}, $w(\bfx_t)=p(\bfx_t \in Y), w(\bfx_s)=p(\bfx_s \in Y)=p(y_s \in Y)$, where $y_s=y(\bfx_s)$. Note the instance weighing in Eq.(\ref{eq:l_gan}) is similar to~\cite{you2019universal}; however the weights are estimated by the order classifier and not by the source classifier.

The total loss for training the ORUDA network is
\begin{equation}
\begin{tabular}{ll}
${\cal L}(F,G_r,G_o,G_d)=$&${\cal L}_{or}(F,G_r)+\gamma ({\cal L}_{ord}(F,G_o)$\\
                &$+ {\cal L}_{dom}(F,G_d)),$
\end{tabular}
\label{eq:total}    
\end{equation}
where $\gamma$ is a hyper-parameter controlling the importance of the order and domain discrimination adversarial losses. The training objective of the minimax game is the following
\begin{equation}
F^{*}, G_r^{*}, G_o^{*} = \arg \min\limits_{F,G_r,G_o} \max\limits_{G_d} {\cal L}(F,G_r,G_o,G_d).
\label{eq:minmax}
\end{equation}
Eq.~(\ref{eq:minmax}) is solved by alternating between optimizing $F,G_r,G_o$ and $G_d$ until the total loss (\ref{eq:total}) converges.

The ORUDA network (Figure~\ref{fig:architecture}) takes as input a batch of source images and a batch of target images. The feature extractor $F$ generates image representations $f=F(\bfx)$ for both batches. The domain discriminator $G_d$ is trained on representations $f$ 
to distinguish between source and target images. The ordinal regressor $G_r$ is trained on a full set of source images. The order classifier $G_o$ takes a pair of source images and their order relationship as input, and pushes the difference of their feature vectors, $F({\bf x}^{'}) - F({\bf x}^{''})$. Source and target image weights are updated over epochs using estimations $p(\bfx_t \in Y)$ and $p(\bfx_s \in Y)$.

\begin{figure}[t]
\centering
\includegraphics[width=0.9\columnwidth]{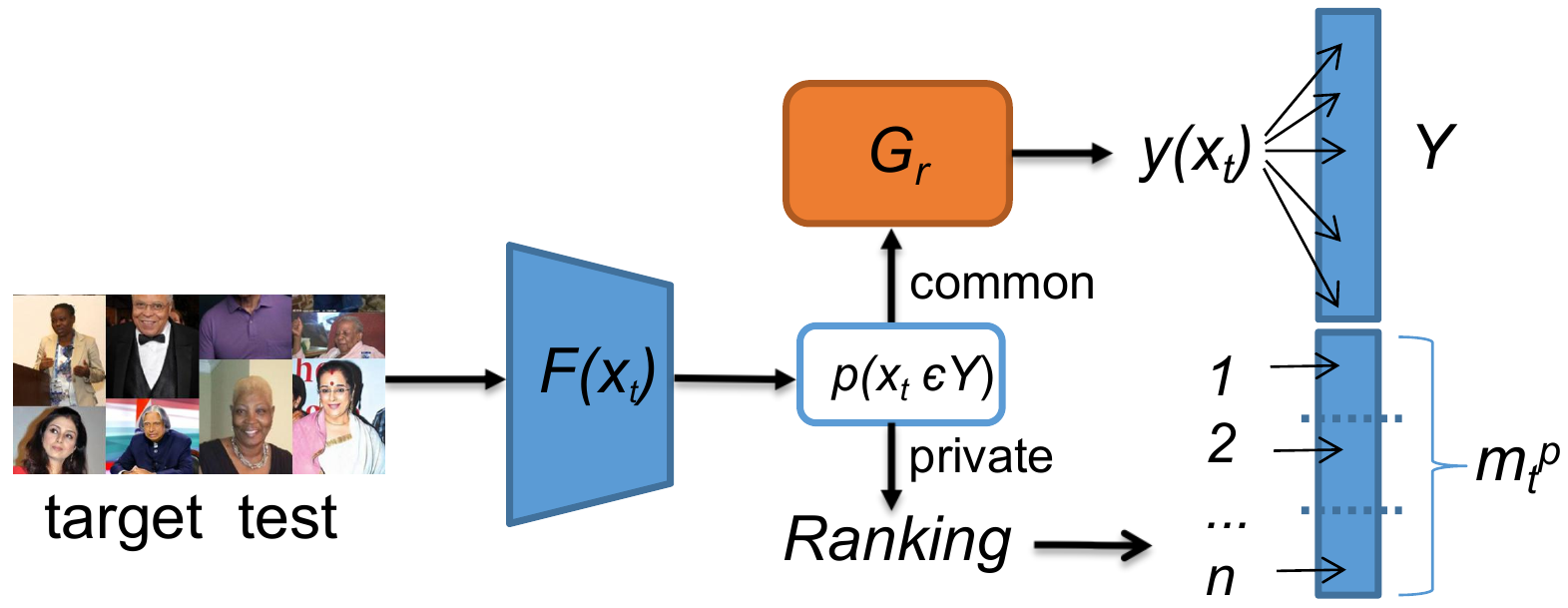}
\caption{Target image evaluation, with classification of  images in $Y$ and the ranking of images in $Y_t^p$.}
\label{fig:arch_test}
\end{figure}
\subsection{Order to ranking to classes}
\label{ssec:ranking}
After training, the ORUDA network 
estimates the boundary between common and private classes.
It marks a source class $y_s$ as common if $p(y_s \in Y)>0.5$ and private otherwise. It marks a target image $\bfx_t$ as common if $p(\bfx_t \in Y)>0.5$ and as private otherwise. For target images marked as common, it uses the ordinal classifier to estimate their classes (see Figure~\ref{fig:arch_test}).
At this stage, classical UDA methods for classification mark all private target images as ``\textit{unknown}'' and terminate. In our case, we can proceed by applying the order model to target private images and classify them using Assumption \ref{as:manifold}b. We proceed in two steps: (i) conversion of pairwise comparisons to rankings, followed by (ii) conversion of rankings to classes.
\paragraph{Image order to image ranking.}
In information retrieval, ranking problems involve a collection of $n$ items and some unknown underlying total ordering of these items. In many applications, one may observe noisy comparisons between items. Examples include matches between teams in a football tournament and consumer’s preference ratings. Given a set of noisy item comparisons, any ranking method tries to find the true underlying ordering of all $n$ items.

The problem of finding approximate rankings based on noisy pairwise comparisons is well studied~\cite{cattelan2012}. In our case, to rank the private target images, we resort to the Bradley–Terry model which is used in multiple applications~\cite{grasshoff08optimal,guo18experimental}. It deals with pairwise comparisons among $n$ images and assumes that there are positive quantities $\pi_i,i=1,\ldots,n$ such that
$p(\bfx_i \prec \bfx_j) = \frac{\pi_i}{\pi_i + \pi_j}$.
Assuming independence of all comparisons, the probability $p_{ij} =p(\bfx_j \prec \bfx_j)$ satisfies the logit model
\begin{equation}
\log \frac{p_{ij}}{1-p_{ij}} = \log \pi_j - \log \pi_i.
\end{equation}
All parameters $\pi_i$ can be estimated by maximum likelihood using standard software for generalized linear models.

Once we detected $n$ private target images, we randomly sample image pairs, apply the order classifier to estimate their relative order and run the Bradley-Terry model to convert the comparisons into a ranking. This private image ranking forms 
an extension of the common class set $Y$ (see Figure~\ref{fig:arch_test}).
The target image ranked 1 is the closest to $Y$, while the target image ranked $n$ is the farthest from $Y$.\footnote{In the case of two target private segments (Figure~\ref{fig:4cases}.b), two separate rankings are necessary.} 
\paragraph{Image ranking to private classes.}
Our model 
classifies the common target images
and ranks the private target images, leading to two different perfmance metrics:
%
\begin{enumerate}
\item Mean square error (MAE) in the common classes $Y$.
\item Ranking error of the private target part $Y_t^p$.  
\end{enumerate}
The first measure is analogous to the classification UDA, where classification error is measured on common classes and ``unknown'' for all images in $Y_t^p$. The ranking error, however, is specific to ORUDA, it estimates how well the order model generalizes to target domain. 

While ranking error is well defined in information retrieval, it is non obvious in OR, where private classes with multiple images per class allow for an {\it exponential number} of totally valid rankings. To simplify the evaluation, we convert the ranking to classes exploiting an assumption of knowing the number of private target classes, $m_t^p=|Y_t^p|$. As the target class thresholds are unknown, we split the full image ranking {\it equally} among $m_t^p$ classes, by simply following their order (see Figure~\ref{fig:arch_test}). The image ranked 1 goes to the first private class while the last ranked image goes to class $m_t^p$. As a result, we can give a true class label to every target image including OS and OSPA cases. We can measure MAE on the entire target set, we denote this extended version {\it e-MAE}. Note that the number of private target classes $m_t^p$ is not used during training nor ranking, it is used for the evaluation only. 
\begin{footnotesize}
\begin{table*}[!ht]
\scriptsize
\begin{center}
\setlength\tabcolsep{2.2pt}
\begin{tabular}{|c||c|c|cc|cc||c|c|cc|cc||c|c|c|cc|} \hline 
Domain pairs
&\multicolumn{6}{c||}{AFAD(A) - UTKFace(U)} &\multicolumn{6}{c||}{CACD(C) - UTKFace(U)}
&\multicolumn{5}{c|}{CACD(C) - AFAD(A)}
\\ \hline
Source 
&A[15-40] &U[1-80] &\multicolumn{2}{c|}{A[15-40]}&\multicolumn{2}{c||}{A[15-40]} 
&C[15-62] &A[15-62]&\multicolumn{2}{c|}{C[15-62]} &\multicolumn{2}{c||}{C[15-62]}
&A[15-40] &C[15-40]&C[15-62]&\multicolumn{2}{c|}{A[15-40]}
\\ \hline
Target 
&U[15-40]&U[15-40]&\multicolumn{2}{c|}{U[1-80]}  &\multicolumn{2}{c||}{U[1-30]} 
&U[15-62]&C[1-80] &\multicolumn{2}{c|}{U[30-80]} &\multicolumn{2}{c||}{U[1-50]} 
&C[15-40] &A[15-40]&A[15-40]&\multicolumn{2}{c|}{C[15-62]}
\\ \hline
Config($\xi$) 
&CS(1.0) &PA(0.325) &\multicolumn{2}{c|}{OS(0.325)}&\multicolumn{2}{c||}{OSPA(0.4)}
&CS(1.0) &PA(0.40) &\multicolumn{2}{c|}{OS(0.40)} &\multicolumn{2}{c||}{OSPA(0.38)}
&CS(1.0) &CS(1.0)  &PA(0.53)&\multicolumn{2}{c|}{OS(0.53)}
\\ \hline
Setup 
&MAE &MAE &MAE &e-MAE &MAE &e-MAE
&MAE &MAE &MAE &e-MAE &MAE &e-MAE
&MAE &MAE &MAE        &MAE &e-MAE
\\ \hline \hline
No adaptation 
& 5.01  & 8.44  & 11.43 & 12.07 & 9.76 &10.16  
& 13.32 & 19.53 & 10.79 & 11.43 & 9.87 &10.45 
&  8.72 &  7.42 & 17.95 & 15.67 & 16.09
\\ \hline
PADA~\cite{cao2018-eccv} 
& 4.72  & 7.71  &  -   & -  & -  & - 
& 10.71 & 14.60 &    - & -  & -  & - 
&  7.71 &  7.11 & 15.71& -  & -
\\
OPDA-BP~\cite{Saito_2018_ECCV}  
& 4.68  & -    & 11.51 & -    & -  & -  
& 10.68 & -    & 10.33 & -    & -  & -
&  7.48 & 6.84 &  -    & 14.51& -
\\
UAN~\cite{you2019universal}    
& 4.74  &  7.67 & 11.39 & -   & 9.75& -      
& 10.67 & 14.74 & 10.39 & -   & 9.75& -  
&  7.57 &  6.73 & 15.75 & 14.79& - 
\\
DANCE~\cite{saito20universal}  
& {\bf 4.53}& 7.36  & 11.23 & -    & 9.57&- 
&10.33      & 14.53 &  9.87 & -    & 9.43& -
&{\bf 7.13} & 6.25  & 15.23 & 14.17& -
\\
ORUDA(ours)   
& 4.70 & {\bf 5.39} & {\bf 9.36}&{\bf 9.65} &{\bf 7.81}&{\bf 7.70}
&{\bf 9.75}&{\bf 11.12}&{\bf 8.91}&{\bf 8.95}&{\bf 7.79}&{\bf 7.88}
& 7.26 &{\bf 6.19}&{\bf 10.31}&{\bf 12.48}&{\bf 12.57}
\\ \hline
Supervised   
& 4.37  & 3.39 & 5.37 & 5.30 & 4.84 & 4.91  
& 4.33  & 5.20 & 6.86 & 6.72 & 5.86 & 5.33
& 6.55  & 3.37 & 6.52 & 6.49 & 6.41
\\ 
\hline
\end{tabular}
\caption{Evaluation of 12 UDA tasks defined on AFAD-UTKFace, CACD-UTKFace and CACD-AFAD domain pairs. MAE values are reported for CS, PA, OS and OSPA tasks; e-MAE values are reported for OS and OSPA tasks.}
\label{tab:eval}
\end{center}
\end{table*}
\end{footnotesize}


\section{Experimental Results} 
\label{sec:evaluation}
\paragraph{Datasets.}
We test our method on three face age estimation datasets. The Asian Face Dataset~\cite{niu16} (AFAD) includes 165,501 faces with age labels 15-40 years ({\small \url{https://github.com/afad-dataset/tarball}}). 
The large scale face (UTKFace) dataset~\cite{zhang17age} includes 16,434 images with the age labels between 1 and 80 years ({\small \url{https://susanqq.github.io/UTKFace}}). 
In the Cross-Age Celebrity dataset (CACD)~\cite{chen16cross}, the total number of images is 159,449 in the age range 14-62 years ({\small \url{http://bcsiriuschen.github.io/CARC}}). 

Images in the CACD dataset are preprocessed such that the faces spanned the whole image with the nose tip being in the center. In UTKFace and AFAD datasets, the centered images were already provided. Each image dataset is randomly divided into 80\% training data and 20\% test data. All images were resized to $128{\times}128{\times}3$ pixels and then randomly cropped to $120{\times} 120{\times}3$ pixels. 
During model evaluation, the $128\times 128\times 3$ face images were center-cropped to a model input size of $120\times 120\times 3$. 

\paragraph{Implementation and Setup.} 
The ORUDA network is implemented in PyTorch. The feature extractor $F$ is fine-tuned on ResNet-34 network. The ordinal regressor $G_r$, order classifier $G_o$ and domain discriminator $G_d$ were all trained from scratch. $G_r$ includes three FC layers with 512, 512 and $m_s$ nodes; $G_o$ is also composed of three FC layers with 512, 512 and 1 nodes. Domain discriminator $G_d$ is similar to one used in the UDA network~\cite{you2019universal}; it includes three FC layers with 1024, 1024 and 1 nodes, interleaved with ReLu layers, the drop-out Bernoulli parameter is 0.5. 

All network components are trained jointly. The network inputs a batch of source images and a batch of target images to fine-tune $F$ and to train $G_r$, $G_o$ and $G_d$. The CL changes the image sampling policy over epochs. Target image and source class weights are updated after each epoch. 

For all experiments, we use the same hyper-parameters, batch-size, and learning rate. We train the network using the Adam optimizer with a learning rate of $lr=10^{-4}$ and a batch size of 64 images. Hyper-parameter $\gamma$ in the ORUDA loss is set to 1.0. The symmetric Butterworth filter is configured with width $w_{\epsilon}=0.9$ and order $n_{\epsilon}=6$. The threshold for the order relationship is $\tau=3$. Training images are shuffled at each epoch before they are fed to the network.

For target images $\bfx_t$ marked as common, the OR classifier predicts their classes as $G_r(F(\bfx_t))$. Target images marked as private are sampled pairwisely for ranking, with at most $l_t=100$ comparisons per image. 
We run each experiment three times and report the average values.

{\bf Baselines.}
We validate the effectiveness of our system 
comparing them to several baselines. 
We build the baselines by disregarding the class order in OR. It permits to apply UDA methods for classification and compare to our method in MAE setup where all private target images are considered as “unknown”. 
We compare the ORUDA method to UAN~\cite{you2019universal} and DANCE~\cite{saito20universal} designed to address all UDA cases in classification; we also add two baselines for the special cases: PADA~\cite{cao2018-eccv} for PA and CS cases, and OPDA-BP~\cite{Saito_2018_ECCV} for OS and CS cases. 
%


\subsection{Evaluation results}
\label{ssec:results}

Table~\ref{tab:eval} presents results for twelve UDA tasks, including five CS, four PA, four OS and three OSPA configurations defined on three domain pairs, AFAD-UTKFace, CACD-UTKFace and CACD-AFAD. 
Commonness $\xi$ varies from 1.0 (CS) to 0.15 (OSPA). 
Table reports MAE evaluation results for SC, PA, OS and OSPA cases and compare them to the baselines. It also reports e-MAE values for OS and OSPA cases. Here, MAE refers to {\it classification setup} described in Section~\ref{ssec:ranking} with the “\textit{unknown}” class for all private target images~\cite{saito20universal,you2019universal}. e-MAE refers to the {\it classification and ranking} setup in OS and OSPA where the private target images are classified using the order classifier followed by ranking. 

The ORUDA method is compared to the source model transfer without adaptation, four UDA baselines for classification and supervised training. {\it Without adaptation}, the OR model (Coral) is trained on source and tested on target data. The {\it supervised OR model} is trained on labeled target data; it aims to provide the lowest error any UDA method would try to achieve on the task.

As the table shows, the smaller values of $\xi$, the larger the performance gap between the supervised and no adaptation cases. Our method tends to halve this gap in many tasks. 
Our method and baseline methods show comparable performance in the CS cases, however our method outperforms by large margin the baselines in PA, OS and OSPA cases.
The e-MAE values are rather close to MAE; this validates Assumption~\ref{as:manifold}.a stated in Section~\ref{ssec:oruda} an supports classification of private target images via ranking. 

 
\paragraph{Ablation study.}
Table~\ref{tab:ablation} evaluates contributions of ORUDA components introduced in Sections~\ref{sec:uda} and~\ref{sec:arch} using AFAD[15-40]$\ra$UTKFace[1-30] OSPA configuration. 

First, we replace the Coral loss with the cross entropy loss used in UDA methods for classification. It makes the class predictions order-inconsistent and leads to performance drop. Second, we test the binary filter $\epsilon(\cdot)$ (\ref{eq:epsilon}) used in the ideal case. 
The binary filter fails to discriminate private source and target images, and marks them as common thus leading to domain misalignment. Instead, using the smoothed filter (SF) $\tilde \epsilon(\cdot)$ helps find the optimal trade-off between false positives and false negatives in common/private split. Curriculum learning (CL) 
helps boost the order model performance, especially on starting epochs. The order classifier achieves 98\% and 89\% on the source train and test sets, respectively. Its accuracy on the target test starts at 50\% and grows to 76\% at the end of training, thanks to domain invariant image representations.


\begin{table}[!ht]
\begin{center}
\begin{tabular}{|l|lll|c|c|} \hline
Model   & Coral      & SF         & CL         & MAE & e-MAE\\ \hline
ORUDA   &            & \checkmark & \checkmark & 8.21& 8.28 \\ 
ORUDA   & \checkmark &            &            & 9.11& 9.07 \\
ORUDA   & \checkmark & \checkmark &            & 8.06& 8.02 \\
\hline
ORUDA   & \checkmark & \checkmark & \checkmark & {\bf 7.81}&{\bf 7.70}  \\ \hline
\end{tabular}
\end{center}
\caption{Ablation study for different ORUDA components on UTKFace-AFAD OSDA task.}
\label{tab:ablation}
\end{table}

\begin{figure}[t]
\centering
\includegraphics[width=0.24\columnwidth]{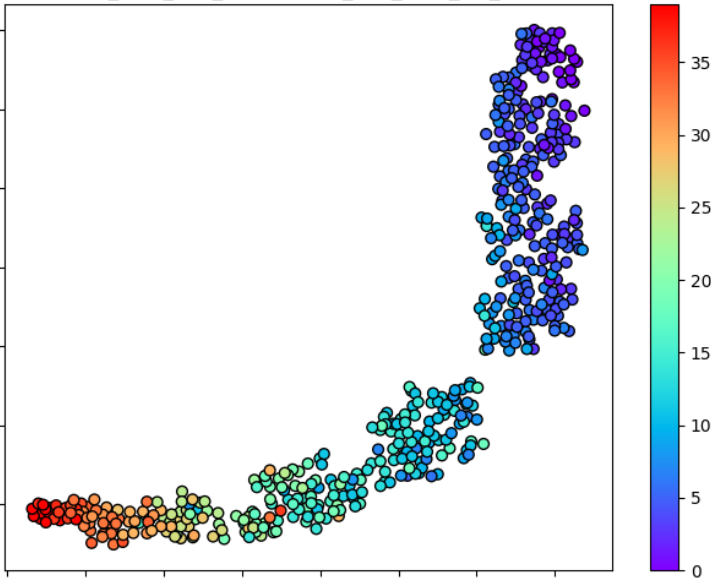}
\includegraphics[width=0.24\columnwidth]{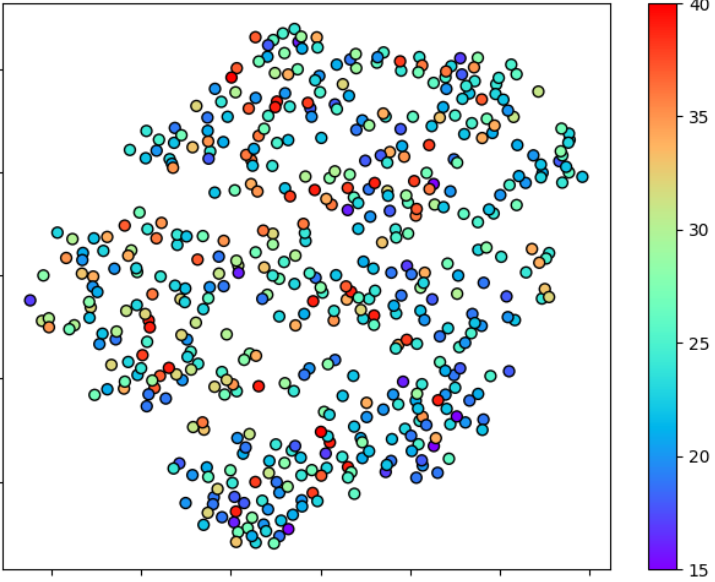}
\includegraphics[width=0.24\columnwidth]{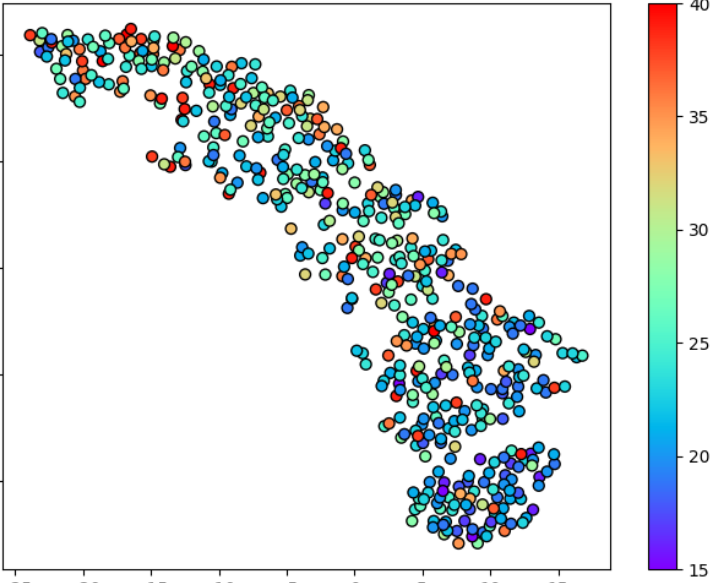}
\includegraphics[width=0.22\columnwidth]{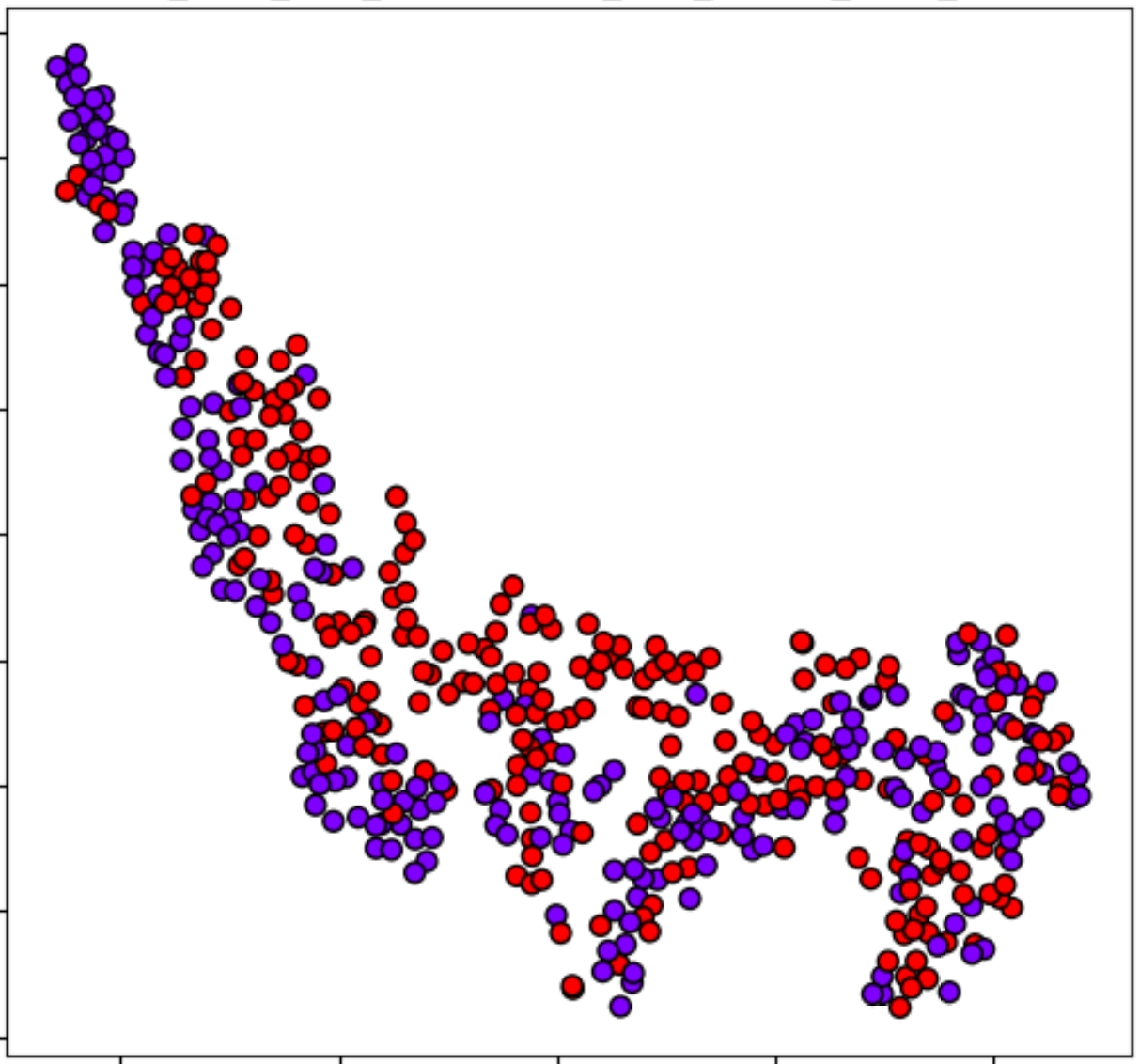}
\caption{TSN-E projections for UTKFace [0-40] to AFAD [15-40] adaptation (left to right): source images; target images before and after adaptation; source and target images with domain colors.}
\label{fig:tsne1}
\end{figure}

\begin{figure}[t]
\centering
\includegraphics[width=0.24\columnwidth]{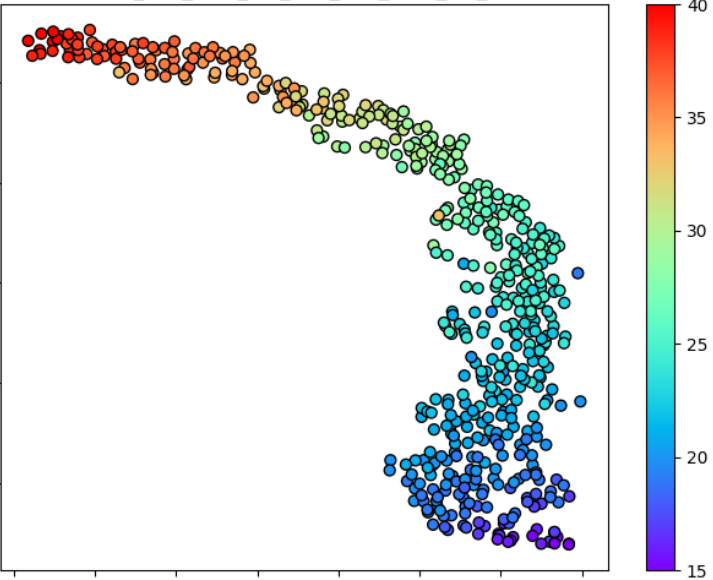}
\includegraphics[width=0.24\columnwidth]{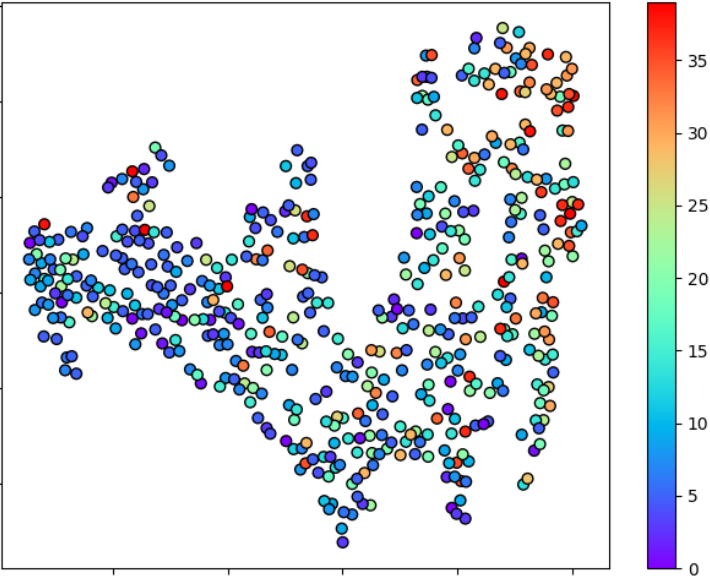}
\includegraphics[width=0.24\columnwidth]{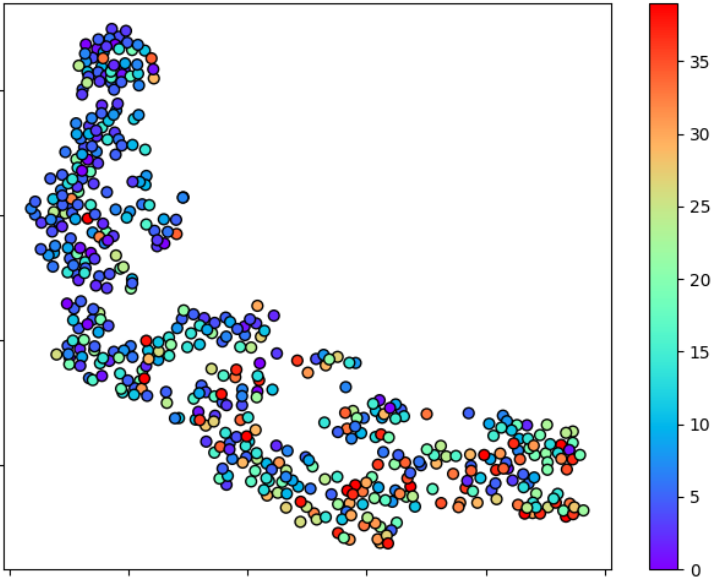}
\includegraphics[width=0.22\columnwidth]{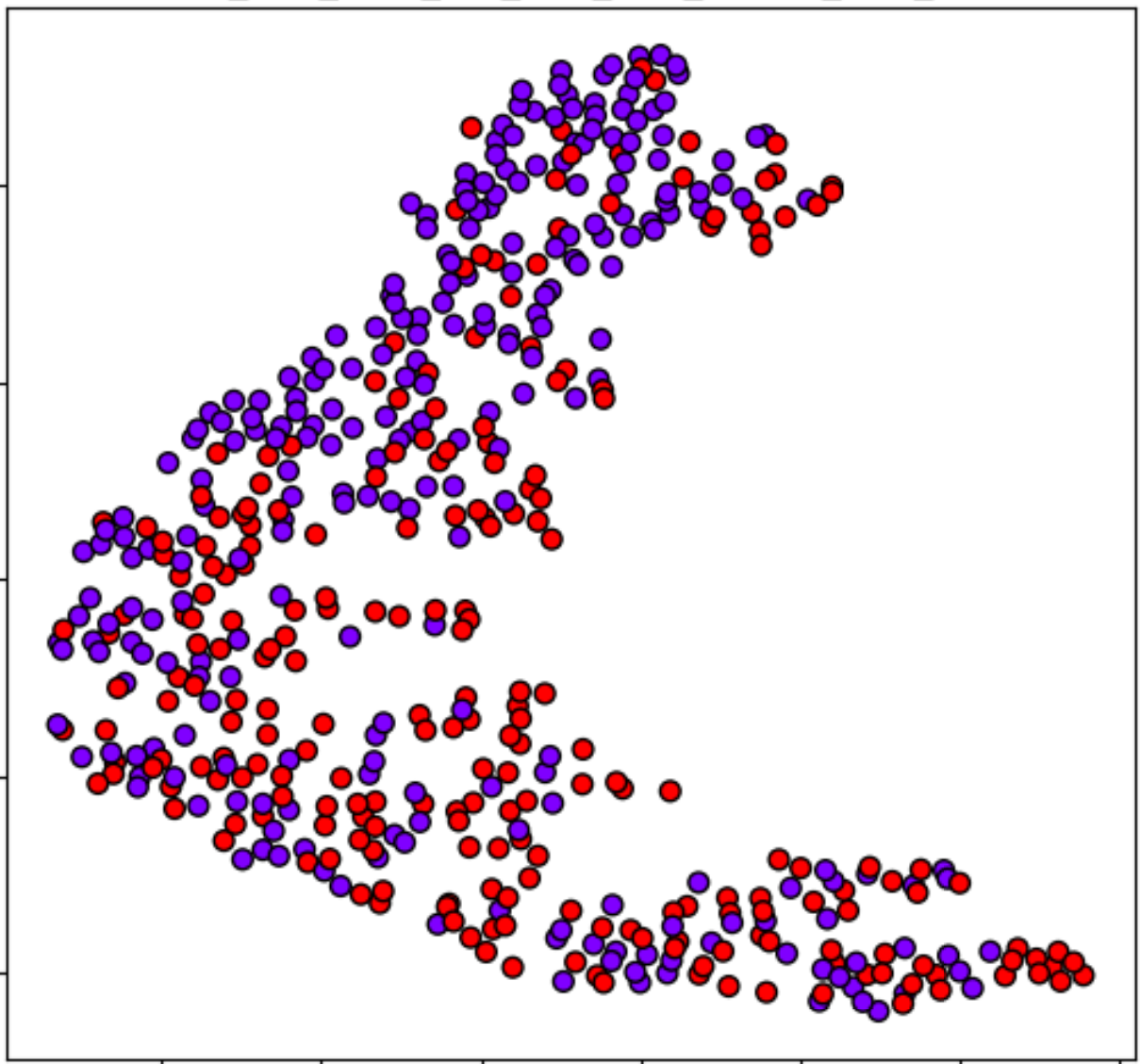}
\caption{TSN-E projections for AFAD [15-40] to UTKFace [0-40] adaptation (left to right): source images; target images before and after adaptation; source and target images with domain colors.}
\label{fig:tsne2}
\end{figure}

\paragraph{TSNE projections.}
Figure~\ref{fig:tsne1} plots four TSN-E projections of image features $f=F(\bfx)$ when adapting UTKFace [0-40] OR model to AFAD [15-40] domain (PA case). The first projection shows a well formed manifold with a perfect class order, obtained by learning the OR model on source images. Second and third projections show target images before and after domain adaptation, with a clearly formed manifold and better ordinal classification. The last projection is on the concatenation of source and target image features after adaptation. Plotted with domain colors (\textcolor{blue}{blue} for source images and \textcolor{red}{red} for target images) it shows how images from the two domains contribute to the common manifold. 
Figure~\ref{fig:tsne2} plots TSN-E projections of image features for the symmetric task of adapting AFAD [15-40] model to UTKFace [1-40] domain (OS case). 

\section{Conclusion}
\label{sec:conclusion}
We addressed the problem of universal domain adaptation in ordinal regression. Instead of the clustering assumption adopted by UDA methods for classification and semantic segmentation, we assume that ordinal data live in a low-dimensional manifold with classes forming a nature order. We proposed a method that complements the OR classifier with an auxiliary task of order learning and  introduced the ORUDA network where the order model can be trained to discriminate between common and private instances, jointly with adversarial domain discrimination. We showed that the order model can expand the natural order to the private target label space on the manifold through ranking. We explained how our model is able to address CS, DA, OS and OSPA configurations. We evaluated the proposed method on a variety of UDA tasks defined on three age estimation datasets and demonstrated its superiority over baselines.


{\small
\bibliographystyle{ieee_fullname}
\bibliography{bib/abbreviations,bib/basic,bib/da,bib/or,bib/cw}
}

\end{document}